\documentclass[10pt, a4paper]{article}
\usepackage{lrec2022} 
\usepackage{multibib}
\newcites{languageresource}{Language Resources}
\usepackage[pdftex]{graphicx}
\usepackage{tikz}
\usetikzlibrary{tikzmark}
\usepackage{multirow}
\usepackage{tabu}
\usepackage{booktabs}
\usepackage{multirow}
\usepackage{colortbl}
\usepackage{subcaption}
\usepackage{tabularx}
\usepackage{amsmath}
\usepackage{amsfonts}
\usepackage{soul}
\usepackage{titlesec}
\titleformat{\section}{\normalfont\large\bfseries\center}{\thesection.}{1em}{}
\titleformat{\subsection}{\normalfont\SmallTitleFont\bfseries\raggedright}{\thesubsection.}{1em}{}
\titleformat{\subsubsection}{\normalfont\normalsize\bfseries\raggedright}{\thesubsubsection.}{1em}{}
\renewcommand\thesection{\arabic{section}}
\renewcommand\thesubsection{\thesection.\arabic{subsection}}
\renewcommand\thesubsubsection{\thesubsection.\arabic{subsubsection}}

\usepackage{epstopdf}
\usepackage[utf8]{inputenc}

\usepackage{xstring}

\usepackage{color}

\title{Domain Mismatch Doesn't Always \\ Prevent Cross-Lingual Transfer Learning}

\name{Daniel Edmiston,$^{1}$ Phillip Keung,$^{1}$ Noah A. Smith$^{2,3}$} 

\address{$^1$Amazon, $^2$University of Washington, $^3$Allen Institute for Artificial Intelligence \\
         danedm@amazon.com, keung@amazon.com, nasmith@cs.washington.edu\\
         }

\abstract{
Cross-lingual transfer learning without labeled target language data or parallel text has been surprisingly effective in zero-shot cross-lingual classification, question answering, unsupervised machine translation, etc. However, some recent publications have claimed that domain mismatch prevents cross-lingual transfer, and their results show that unsupervised bilingual lexicon induction (UBLI) and unsupervised neural machine translation (UNMT) do not work well when the underlying monolingual corpora come from different domains (e.g., French text from Wikipedia but English text from UN proceedings). In this work, we show that a simple initialization regimen can overcome much of the effect of domain mismatch in cross-lingual transfer. We pre-train word and contextual embeddings on the concatenated domain-mismatched corpora, and use these as initializations for three tasks: MUSE UBLI, UN Parallel UNMT, and the SemEval 2017 cross-lingual word similarity task. In all cases, our results challenge the conclusions of prior work by showing that proper initialization can recover a large portion of the losses incurred by domain mismatch.
 \\ \newline  }

\begin{document}

\maketitleabstract

\section{Introduction}

Zero-shot cross-lingual transfer via representation learning has been studied in many recent works spanning a variety of tasks: cross-lingual text classification and named entity recognition \cite{DevlinEtAl2018}, unsupervised neural machine translation 
\cite{LampleEtAl2018,ArtetxeEtAl2018} and unsupervised bilingual lexicon induction \cite{muse,ZhangEtAl2017}, among others. Cross-lingual transfer techniques typically assume that the source and target text come from the same domain (e.g., English and French Wikipedia for UNMT), but many recent papers have reported issues in the domain-mismatched setting (e.g., English Europarl and French Wikipedia).

Particularly, domain mismatch has been shown to have a pernicious effect on UBLI, and has been labeled a ``core limitation'' \cite{Sogaard}, with word embeddings pre-trained on domain-mismatched corpora showing markedly degraded scores. In the case of UNMT, mismatched domains between source and target training data have also been shown to cause large reductions in BLEU scores \cite{marchisio2020does}. The results in Table \ref{UNMT UBLI Results} illustrate the severity of the problem.

\begin{figure} 
    \centering
    \begin{tabular}{c|cc}
         & Language A & Language B \\
         \hline 
        \multirow{2}{*}{Domain 1} &  & \multirow{2}{*}{\cellcolor{lightgray} } \\
         &  & \cellcolor{lightgray} \\
        \multirow{2}{*}{Domain 2} & \multirow{2}{*}{\cellcolor{lightgray} } &  \\
         & \cellcolor{lightgray} &  \\
    \end{tabular}
    \begin{tikzpicture}[overlay, remember picture]
    \node (LangADom1) at (-3.3,.35) {};
    \node (LangADom2) at (-3.3,-.7) {};
    \node (LangBDom1) at (-1.1,.35) {};
    \node (LangBDom2) at (-1.1,-.7) {};
    \draw[draw=blue, ->, line width=.33mm] (LangADom2) edge (LangADom1);
    \draw[draw=blue, ->, line width=.33mm] (LangADom2) edge (LangBDom2);
    \draw[draw=blue, ->, line width=.33mm] (LangBDom1) edge (LangADom1);
    \draw[draw=blue, ->, line width=.33mm] (LangBDom1) edge (LangBDom2);
    \end{tikzpicture}
    \caption{In our zero-shot experiments, training corpora for each language belong to different domains (grey cells). We show joint pre-training induces cross-domain, cross-lingual transfer (arrows) even when no data exists in the double-crossed scenario (white cells).}
    \label{Scenario schematic}
\end{figure}
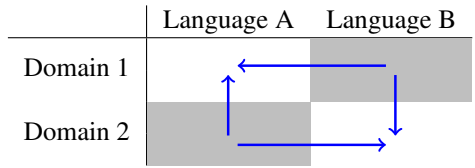

\begin{table*}[t!]
\footnotesize
	\begin{minipage}{1.0\linewidth}
		\centering
		\begin{tabu}{lllrr}
			\toprule
			 & \textbf{Language Pair} & \textbf{Source/Target Domain} & \textbf{Score} & $\Delta$ \textbf{Mismatch vs. Match} \\
			\midrule 
			\emph{UBLI} &&& \emph{Accuracy@1} & \\
			 \midrule 
			 \multirow{4}{*}{\newcite{Sogaard}} & \multirow{1}{*}{English-Spanish} & Europarl/Europarl & 61.0 &  \\
			 & & Europarl/Wikipedia & 0.1 & --60.9 \\
			  & \multirow{1}{*}{English-Hungarian} & Wikipedia/Wikipedia & 6.7 &  \\
			  & & Wikipedia/Europarl & 0.1 & --6.6 \\
			 \midrule 
			 \emph{UNMT} & & & \emph{BLEU} & \\
			\midrule 
			\multirow{4}{*}{\newcite{marchisio2020does}} & \multirow{1}{*}{French-English} & UN Parallel/UN Parallel  & 27.6 &  \\
			 & & UN Parallel/Common Crawl& 3.3 & --24.3 \\
			 &\multirow{1}{*}{Russian-English} & UN Parallel/UN Parallel & 23.7 &  \\
			 & & UN Parallel/Common Crawl & 0.7 & --23.0 \\
			\bottomrule
		\end{tabu}
	\end{minipage}
	\caption{Unsupervised bilingual lexicon induction (UBLI) and neural machine translation (UNMT) results from some previous papers. When the domains of the monolingual text are mismatched, UBLI and UNMT yield retrieval accuracy and BLEU scores very close to 0.}
	\label{UNMT UBLI Results}
\end{table*}

In this work, we show that cross-lingual transfer can occur \emph{even when there is no overlap between the domains in the same language and no overlap between the languages in the same domain} (Figure \ref{Scenario schematic}). Earlier work such as mBERT \cite{DevlinEtAl2018} and XLM \cite{xlm} demonstrated that pre-training contextual embeddings on concatenated multilingual Wikipedia text induces cross-lingual transfer effects. We extend these findings to the domain-mismatched case, where we pre-train our embeddings on concatenated multilingual domain-mismatched text. We compare the effect of initializing with and without joint pre-training for three cross-lingual tasks: MUSE BLI \cite{muse}, UN Parallel MT \cite{un}, and SemEval 17 cross-lingual word similarity \cite{CamachoEtAl2017}. Contrary to the findings on UBLI and UNMT from recent publications, we find that the availability of domain-matched corpora is not a prerequisite for effective cross-lingual transfer, since the domain mismatch issue can be mitigated by using an appropriate initialization.

\section{Unsupervised BLI Experiments}

\subsection{Background}
Bilingual lexicon induction refers to a word translation task, with modern methods relying on retrieval in a continuous space shared by both source and target embeddings. BLI has successfully used small seed dictionaries as a form of cross-lingual signal \cite{MikolovEtAl2013,duong-etal-2016-learning}, but recent unsupervised alternatives have proven competitive with supervised approaches \cite{ArtetxeEtAl2018Robust,HeymanEtAl2019}, with methods based on adversarial learning \cite{ZhangEtAl2017,muse} and point-cloud matching \cite{hoshen-wolf-2018-non}.

We use the MUSE model of \newcite{muse} for all UBLI experiments, with all experiments conducted on AWS p3.16xlarge hosts. MUSE uses an adversarial objective \cite{gan} to learn a transformation from the word embedding space of the source language to that of the target language, along with a discriminator to distinguish transformed source embeddings from target embeddings. Word translation is achieved using margin-based nearest-neighbors to retrieve target embeddings from transformed source embeddings, and evaluated using the test sets provided with MUSE.

In UBLI with MUSE, word embeddings for the source and target languages are pre-trained independently (e.g., French and English word embeddings are trained separately on Wikipedia text). When the domains are mismatched (e.g., Wikipedia and UN), UBLI retrieval accuracy has been shown to suffer greatly \cite{Sogaard}, as mentioned above.

We compare the standard way of pre-training for MUSE UBLI---pre-training word embeddings separately for each language---with joint pre-training, where we train multilingual word embeddings on concatenated domain-mismatched corpora \cite{LampleEtAl2018}. Note that this is a simple form of multilingual joint pre-training, and does not include any \textit{post-hoc} processing steps (cf. \newcite{WangEtAl2020}, who perform a vocabulary reallocation step to eliminate spurious anchors in the shared embedding space\footnote{Spurious anchors are embeddings in the shared space which result from forms which appear in both languages, but which have different meanings in each language, e.g. \textit{coin} in English and French. In French \textit{coin} means \textit{corner}, and therefore \textit{coin} should not map to a single vector in the shared embedding space.}).

We study the following language pairs in both directions: English-French, English-Spanish, and English-Russian.\footnote{We lemmatize all Russian data for the UBLI experiments with the \textit{pymorphy2} \cite{Korobov15} morphological analyzer, abstracting out challenges posed by morphologically rich languages like Russian.} For each experiment, we initialize embeddings via \textit{fastText} \cite{fasttext} using Wiki and UN corpora\footnote{We used Wiki dumps from June 2020 and UN corpus v1.0 \cite{un}. To address the disparity in corpus sizes, we sampled 5M lines from each for training. Tokenization was done with Moses \cite{KoehnEtAl2007}.} and perform grid search over MUSE hyperparameters, reporting scores for the configuration with the highest CSLS (cross-domain similarity local scaling) score, which is an unsupervised metric discussed in \newcite{muse}.\footnote{The unsupervised CSLS scores are computed using only the training corpora.} We optimize CSLS score for four random seeds (123, 456, 789, and 321), three choices of iterations for Procrustes refinement (1, 3, and 5), and three choices of epochs (1, 3, and 5).

\begin{table*}
\begin{minipage}{1.0\linewidth}
	\centering
\begin{tabu}{lrrrrrr}
\toprule
\textbf{Source Domain-Target Domain} & \textbf{Es-En} & \textbf{En-Es} & \textbf{Fr-En} & \textbf{En-Fr} & \textbf{Ru*-En} & \textbf{En-Ru*} \\
  \midrule 
  \textit{Matched Domain} \\
  \midrule 
Wiki - Wiki & 81.8 & 82.5 & 81.3 & 82.2 & 59.5 & 64.0 \\
UN - UN & 68.7 & 70.8 & 74.2 & 75.2 & 55.2 & 56.7 \\
\midrule
 \textit{Mismatched Domain} \\
 \midrule 
Wiki - UN & 0.1 & 0.2 & 35.2 & 33.8 & 16.6  & 0.1 \\
Wiki - UN w/ Joint Pre-training & 65.2 & 56.9 & 68.3 & 54.4 & 28.0 & 20.1 \\
$\Delta$ & +65.1 & +56.7 & +33.1 & +20.6 & +11.4 & +20.0 \\

\bottomrule
\end{tabu}
\end{minipage}
\caption{Retrieval accuracy@1 for UBLI across language pairs on MUSE test dictionaries. \emph{Joint pre-training} uses word embeddings that are jointly pre-trained on concatenated source Wiki and target UN corpora. Ru* denotes lemmatized Russian. }
\label{UBLI Results}
\end{table*}

\subsection{Results}

We present our retrieval accuracies at 1 for UBLI experiments in Table \ref{UBLI Results}. As expected, MUSE works well with domain-matched corpora, while our domain-mismatched experiments show large degradations relative to matched domain baselines in all cases. In particular, scores for Es-En, En-Es, and En-Ru all fall to near 0.0, showing that cross-lingual transfer has failed in these cases. However, in all cases, joint pre-training recovers a large portion of the losses incurred by mismatched corpora, showing cross-lingual transfer is still possible, contrary to the conclusions drawn in \newcite{Sogaard} and \newcite{VulicEtAl2019}. 

\subsection{The Role of Identical Words in UBLI Performance}

It is important to note that a large proportion of word pairs in the MUSE test dictionaries are identical (e.g., \textit{Paris}-\textit{Paris} in Fr-En), and joint-training is able to take advantage of identical spellings, since words with the same spelling will always have the same word embedding \cite{LampleEtAl2018}. In Table \ref{Comparison with copying baseline}, we show the performance of the `copying baseline', which simply treats each word as its own translation. This baseline is surprisingly strong; its accuracy at 1 exceeds 40\% for English-French in both directions.

\newcite{Sogaard} explicitly use identical words to create a seed dictionary to improve performance in the cross-domain scenario. While \newcite{Sogaard}'s best reported score for En-Es with the seed dictionary approach actually falls below the simple copying baseline at 25.5\%, we show that the joint pre-training approach yielded 56.9\%. A comparison between results from the two approaches and the copying baseline are shown in Table \ref{Comparison with copying baseline}.

\begin{table*}[t!]
\footnotesize
	\begin{minipage}{1.0\linewidth}
		\centering
		\begin{tabu}{lcccccc}
			\toprule
			 & \textbf{Language Pair} & \textbf{Src/Tgt Domain} & \textbf{Acc@1} & \textbf{Copying Baseline} & \textbf{$\Delta$} & \textbf{STDM} \\
			\midrule 
			 \multirow{3}{*}{\newcite{Sogaard}} & \multirow{1}{*}{En-Es} & & 25.5 & 32.8 & -7.3 & 0.27 \\
			 & \multirow{1}{*}{En-Fi}  &Wiki/Europarl & 10.1 & 28.8 & -18.7 & 0.20 \\
			  & \multirow{1}{*}{En-Hu} & & 9.2 & 29.6 & -20.4 & 0.20 \\
			 \midrule 
			\multirow{6}{*}{Joint pre-training (this paper)}
			 & En-Es & \multirow{6}{*}{Wiki/UN}  & 56.9 & 32.8 & +24.1 & 0.25 \\
			 & Es-En & & 65.2 & 29.7 & +35.5 & 0.33 \\
			 & En-Fr & & 54.4 & 41.3 & +13.1 & 0.28 \\
			 & Fr-En & & 68.3 & 42.9 & +25.4 & 0.33 \\
			 & En-Ru & & 20.1 & 3.9 & +16.2 & 0.25 \\
			 & Ru-En & & 28.0 & 0.0 & +28.0 & 0.33 \\
			\bottomrule
		\end{tabu}
	\end{minipage}
	\caption{Accuracies based on the method of \newcite{Sogaard} and joint pre-training versus a simple copying baseline. $\Delta$ refers to difference between copying baseline and Acc@1 score for each method. STDM refers to the Source-Target Domain Mismatch score for each corpus and language pair, as described in \newcite{ShenEtAl2021}.} 
	\label{Comparison with copying baseline}
\end{table*}

\subsection{The Source-Target Domain Mismatch (STDM) Score}\label{Measuring STDM}

The considerable improvement of joint-training over the unsupervised seed dictionary method as detailed in Table \ref{Comparison with copying baseline} could be the result of the relative distances between source and target domains rather than of the differing techniques. For instance, if the Wikipedia and UN corpora were more similar than the Wikipedia and Europarl corpora, domain mismatch would be more pronounced in the Wikipedia-Europarl case and lower scores would be expected. The recently proposed Source-Target Domain Mismatch (STDM) score of \newcite{ShenEtAl2021} provides a means of measuring domain similarity between corpora, and we show that the Europarl and UN corpora are not dramatically different from the Wikipedia corpora in all cases. This suggests that the disparity between the results is attributable to the different methods rather than the relative similarity between domains.

The STDM score is computed in the following way. Let $\mathcal{A}=concat(A_{1}, A_{2})$, or the concatenation of two corpora $A_{1}$ and $A_{2}$, where $A_{1}$ and $A_{2}$ are the corpora we wish to compare and which consist of $n$ and $m$ documents respectively. Then let $\mathcal{A}_{\text{\textit{tfidf}}}\in\mathbb{R}^{(n+m)\times\vert V\vert}$ be the result of applying TF-IDF \cite{Sparck1972} to the concatenated corpora, thus giving a matrix whose top $n$ rows are representations of the documents from $A_{1}$ and the bottom $m$ rows are representations of the documents from $A_{2}$. Then the (truncated) SVD decomposition of $\mathcal{A}_{\text{\textit{tfidf}}}$ is $\bar{\mathcal{A}}_{\text{\textit{tfidf}}}=USV=(U\sqrt{S})(\sqrt{S}V)=\bar{U}\bar{V}$, where $\bar{U}$ contains topic representations of the corpora's documents, again with the first $n$ rows ($\bar{U}_{1}$) being the representations of corpus $A_{1}$ and the bottom $m$ rows ($\bar{U}_{2}$) the representations of corpus $A_{2}$. Then define $s_{A,B}$ as in Equation \ref{s score for STDM}: 

\begin{equation}\label{s score for STDM}
    s_{A,B} = \frac{1}{n \cdot m} \sum_{i=1}^{n} \sum_{j=1}^{m} (\bar{U}_{A}\bar{U}_{B}^{\top})_{i,j}
\end{equation}

$s_{A,B}$ then measures the average similarity of documents between corpus $A$ and corpus $B$. Given $s_{A,B}$ for each combination of corpora, then the STDM score is defined as in Equation \ref{STDM score}.

\begin{equation}\label{STDM score}
    \text{\textit{STDM}} = \frac{s_{1,2} + s_{2,1}}{s_{1,1}+s_{2,2}}
\end{equation}

The STDM score, which in practice ranges from 0 (completely dissimilar) to 1 (identical), thus uses latent semantic analysis (LSA) \cite{DumaisEtAl1988} on a combined corpus to derive topic representations of the individual corpora, using the intuition that similar corpora will have similar documents. We use this score to quantify the similarity between mismatched corpora and rule out relative domain divergence as a causal factor in the disparity between our scores and those of \newcite{Sogaard}.\footnote{Note that a corpus similarity score based on TF-IDF cannot compare corpora from different languages directly. \newcite{ShenEtAl2021} work around this issue by comparing the corpora in the target language. For example, to quantify the domain mismatch in an English-Wikipedia to Spanish-Europarl experiment, the corpora of comparison would be the Spanish Wikipedia and the Spanish Europarl, since Spanish and English Wikipedia would cover similar topics.}

The STDM scores for the mismatched corpora for each experiment are found in the STDM column of Table \ref{Comparison with copying baseline}. The STDM comparisons between corpora show that, while on average the Wikipedia and UN corpora are more similar than the Wikipedia and Europarl corpora, this difference is small and unlikely to account for the large disparity in results. On the one point of direct comparison, namely UBLI from English into Spanish, the Wikipedia and Europarl corpora are shown to be more similar (STDM=0.27) than the Wikipedia and UN corpora (STDM=0.25), yet joint-training on the more dissimilar corpora still produces better results (56.9\% vs. 25.5\%).  

\subsection{On Domain Mismatch in UBLI}

We emphasize three points from the results of these experiments. Firstly, in many cases, initialization by a simple joint-training regimen can largely overcome the deleterious effects of domain mismatch for the UBLI task, challenging the conclusions of \newcite{Sogaard} and \newcite{VulicEtAl2019}. This is seen primarily in experiments involving closely related languages (English-Spanish, English-French), where mismatched domain experiments run with joint-training initialization approach the scores of the matched domain experiments.

Secondly, while still beneficial, joint-training initialization is less effective on distantly related languages, such as English-Russian. Improvements from joint-training are considerable, but scores still fall below the matched domain experiments, suggesting that this method does not fully solve domain mismatch in all cases. 

Lastly, task-agnostic joint-training initialization performs favorably when compared against the identical-word seed dictionary method of \newcite{Sogaard} in terms of ameliorating the effects of domain mismatch, as shown by comparison of each method against a copying baseline.

\begin{table*}[t!]
	\tiny
	\centering
	\resizebox{1.0\textwidth}{!}{%
	\begin{tabu}{lllrrrrrrrrrr}
		\toprule 
		\textbf{Monolingual Data} & \textbf{Task} &  & \textbf{Epoch 1} & \textbf{2} & \textbf{3} & \textbf{4} & \textbf{5} & \textbf{6} & \textbf{7} & \textbf{8} & \textbf{9} & \textbf{10} \\
		\midrule 
		 &  & UNMT Baseline & 3.37 & 3.65 & 4.01 & 4.40& 4.30 & 4.22 & 4.17 & 3.73 & 4.16 & 3.99 \\
		En Wiki, Fr UN & En UN $\rightarrow$ Fr UN & w$/$ Joint Pre-train. & 22.76 & 24.61& 24.83 & 25.19 & 24.98 & 25.09 & 24.95 & 24.95 & 24.70 & \textbf{25.45} \\
		&& $\Delta$ & +19.39 & +20.96 & +20.82 & +20.79 & +20.68 & +20.87 & +20.78 & +21.22 & +20.54 & \textbf{+21.46} \\
		\midrule 
		 &  & UNMT Baseline & 4.11 & 4.01 & 3.99 & 4.17 & 4.35 & 4.41 & 4.65 & 4.87 & 5.21 & 5.39 \\
		Fr Wiki, En UN & Fr UN $\rightarrow$ En UN & w$/$ Joint Pre-train. & 19.78 & 20.20 & \textbf{20.91} & 19.94 & 19.62 & 19.86 & 19.68 & 19.62 & 20.54 & 19.80 \\
		&& $\Delta$ & +15.67 & +16.19 & \textbf{+16.92} & +15.77 & +15.27 & +15.45 & +15.03 & +14.75 & +15.33 & +14.41 \\
		\midrule 
		 &  & UNMT Baseline & 1.50 & 1.04 & 1.45 & 1.46 & 1.59 & 1.35 & 1.29 & 1.32 & 1.37 & 1.37 \\
		En Wiki, Ru UN & En UN $\rightarrow$ Ru UN & w$/$ Joint Pre-train. & \textbf{10.83} & 9.14 & 9.62 & 8.96 & 8.82 & 9.79 & 10.09 & 10.37 & 9.82 & 9.71 \\
		&& $\Delta$ & \textbf{+9.33} & +8.10 & +8.17 & +7.50 & +7.23 & +8.44 & +8.80 & +9.05 & +8.45 & +8.34 \\
		\midrule 
		 & & UNMT Baseline & 1.51 & 1.27 & 1.52 & 1.53 & 1.31 & 1.22 & 1.40 & 1.28 & 1.23 & 1.22 \\
		Ru Wiki, En UN & Ru UN $\rightarrow$ En UN & w$/$ Joint Pre-train. & 7.71 & \textbf{7.95} & 7.82 & 7.43 & 7.46 & 7.49 & 6.45 & 6.85 & 6.60 & 6.77 \\
		&& $\Delta$ & +6.20 & \textbf{+6.68} & +6.30 & +5.90 & +6.15 & +6.27 & +5.05 & +5.57 & +5.37 & +5.55 \\
		\bottomrule 
	\end{tabu}%
	}
\caption{UNMT BLEU scores with and without joint pre-training on the UN Corpus development sets. (Highest BLEU scores in each row are bolded.) Initializing with jointly pre-trained contextual embeddings yields up to a 21 point gain in BLEU. BLEU score improvements are always larger when the monolingual pre-training data contains target language text in the UN domain.}
\label{unmt_results}
\end{table*}

\section{Unsupervised NMT Experiments}

\subsection{Task Description}

Unsupervised NMT systems address the paucity of available parallel data for most language pairs, relying only on monolingual data from the source and target languages. The models of \newcite{LampleEtAl2018} and \newcite{ArtetxeEtAl2018} are representative, each employing encoder-decoder architectures with weight-sharing between languages. Models are trained via the dual tasks of sentence reconstruction and back-translation \cite{SennrichEtAl2016}.\footnote{See \newcite{WuEtAl2019} and \newcite{keung_bitext} for alternatives to back-translation.} Follow-up work has incorporated statistical machine translation (SMT)  systems \cite{smt}; \newcite{ArtetxeEtAl2019} and \newcite{MarieFujita2018} use unsupervised SMT systems to initialize UNMT systems, while \newcite{RenEtAl2019} incorporate SMT as a form of posterior regularization.

For all UNMT experiments, we adopt the encoder-decoder model of \newcite{lample2018phrase}, a sequence-to-sequence model with 6 transformer layers for both the encoder and decoder, and use the implementation provided by the authors.\footnote{https://github.com/facebookresearch/XLM} We study the English-French and English-Russian language pairs in both directions, training all models for ten epochs. We used 5 million sentences per language in the monolingual data used for UNMT training. We trained two English-French and two English-Russian UNMT models on the following sets of 10 million sentences: (En Wiki, Fr UN), (En UN, Fr Wiki), (En Wiki, Ru UN), and (En UN, Ru Wiki).

In a manner similar to our UBLI experiments, we compare UNMT performance with and without jointly pre-trained contextual embeddings. In the baseline system, we follow the UNMT approach outlined in \newcite{xlm}, where the encoder and decoder are initialized with a contextual embedding pre-trained on Wikipedia text only. In the jointly pre-trained case, the encoder and decoder are initialized with a contextual embedding that was pre-trained on a mix of Wikipedia and UN text.

\subsection{Results}

In Tables \ref{unmt_results} and \ref{unmt_results_no_un}, we show the difference in the UN development BLEU scores between a UNMT system with and without joint pre-training. Table \ref{unmt_results} shows results for experiments in which the monolingual data contains UN text in the target language, while Table \ref{unmt_results_no_un} shows results for experiments in which it does not. We report results for the first 10 epochs (where one epoch consists of 200k sentences), and the tables show BLEU scores as training progresses. 

As we saw in previous work with domain mismatch (Table \ref{UNMT UBLI Results}), the baseline system without joint pre-training fails to learn to translate; BLEU scores never exceed 6 for any language pair.\footnote{\newcite{lample2018phrase} show that in a domain-matched Fr-En setting, the baseline UNMT system can achieve $>24$ BLEU on WMT'14.} However, with joint pre-training, we observe as much as 21 points of BLEU improvement. An additional 10 epochs of training (not shown) do not yield a BLEU improvement for the baseline.



The difference between results in Tables \ref{unmt_results} and \ref{unmt_results_no_un} shows that UNMT performance depends strongly on whether the monolingual data contains UN text in the target language. For example, Fr UN $\rightarrow$ En UN BLEU scores are higher when the monolingual data is (Fr WIKI, En UN) and lower when it is (Fr UN, En Wiki). 

\begin{table*}[t!]
	\tiny
	\centering
	\resizebox{1.0\textwidth}{!}{%
	\begin{tabu}{lllrrrrrrrrrr}
		\toprule 
		\textbf{Monolingual Data} & \textbf{Task} &  & \textbf{Epoch 1} & \textbf{2} & \textbf{3} & \textbf{4} & \textbf{5} & \textbf{6} & \textbf{7} & \textbf{8} & \textbf{9} & \textbf{10} \\
		\midrule 
		&  & UNMT Baseline & 3.94 & 3.84 & 3.75 & 3.77 & 3.89 & 3.87 & 3.87 & 4.12 & 4.21 & 4.40 \\
		En UN, Fr Wiki & En UN $\rightarrow$ Fr UN &w$/$ Joint Pre-train. & 11.78 & 12.50 & \textbf{12.76} & 11.73 & 11.90 & 11.65 & 11.20 & 11.24 & 11.62 & 11.59 \\
		&& $\Delta$ & +7.84 & +8.66 & \textbf{+9.01} & +7.96 & +8.01 & +7.78 & +7.33 & +7.12 & +7.41 & +7.19 \\
		\midrule 
		&  & UNMT Baseline & 2.66 & 3.79 & 3.70 & 4.32 & 4.11 & 4.33 & 4.11 & 4.14 & 4.18 & 4.04 \\
		Fr UN, En Wiki & Fr UN $\rightarrow$ En UN &w$/$ Joint Pre-train. & 13.48 & 13.82 & \textbf{14.19} & 13.40 & 13.70 & 13.39 & 13.34 & 13.38 & 12.74 & 13.65 \\
		&& $\Delta$ & +10.82 & +10.03 & \textbf{+10.49} & +9.08 & +9.59 & +9.06 & +9.23 & +9.24 & +8.56 & +9.61 \\
		\midrule 
		&  & UNMT Baseline & 1.03 & 1.09 & 1.19 & 0.89 & 0.79 & 0.98 & 0.79 & 0.79 & 0.80 & 0.81 \\
		En UN, Ru Wiki & En UN $\rightarrow$ Ru UN &w$/$ Joint Pre-train. & 4.47 & 4.41 & 4.44 & 4.40 & \textbf{4.52} & 4.38 & 4.10 & 4.36 & 4.11 & 3.99 \\
		&& $\Delta$ & +3.44 & +3.32 & +3.25 & +3.51 & \textbf{+3.73} & +3.40 & +3.31 & +3.57 & +3.31 & +3.18 \\
		\midrule 
		& & UNMT Baseline & 1.24 & 0.97 & 0.99 & 1.29 & 1.49 & 1.47 & 1.41 & 1.64 & 1.57 & 1.42 \\
		Ru UN, En Wiki & Ru UN $\rightarrow$ En UN &w$/$ Joint Pre-train. & \textbf{7.63} & 7.57 & 6.93 & 6.98 & 7.31 & 6.89 & 6.70 & 7.11 & 7.02 & 6.56 \\
		&& $\Delta$ & \textbf{+6.39} & +6.60 & +5.94 & +5.69 & +5.82 & +5.42 & +5.29 & +5.47 & +5.45 & +5.14 \\
		\bottomrule 
	\end{tabu}%
	}
\caption{UNMT BLEU scores when the monolingual data \emph{doesn't} contain target language UN text, in contrast to Table \ref{unmt_results}.}
\label{unmt_results_no_un}
\end{table*}

Furthermore, while we did not perform matched domain UNMT experiments for comparison due to computational constraints, we can take the French-English and Russian-English experiments from \newcite{marchisio2020does} (see again Table \ref{UNMT UBLI Results}) to establish the extent to which joint-training ameliorates domain mismatch. \newcite{marchisio2020does} report a BLEU score of 27.6 for the Fr UN - En UN experiment, and a BLEU score of 23.7 for the Ru UN - En UN experiment. In our best domain-mismatched experiment for the Fr UN - En UN task, joint-training resulted in a BLEU score of 20.91, making up most of the losses incurred by domain mismatch. However, in our best domain-mismatched experiment for the Ru UN - En UN task, the BLEU score was only 7.95, which is well below the domain matched score of 23.7 reported in \newcite{marchisio2020does}. Thus, similar to the UBLI experiments above, joint-training results in a pronounced recovery of cross-lingual transfer when the language pair is similar, but results in only modest gains for more distant language pairs.

\section{Cross-lingual Semantic Word Similarity Experiments} 

\subsection{Task Description}

In addition to investigating the UBLI and UNMT tasks, we also examine cross-lingual transfer via word similarity tasks, and in doing so show that joint-training via concatenation is useful generally, even for non-translation related tasks. The semantic word similarity task consists of evaluating pairs of words (e.g. WS353) \cite{Finkelstein2001EtAl} via a similarity metric (e.g., cosine similarity) on their embeddings, and comparing these scores with human judgments. The SemEval 17 cross-lingual semantic word similarity task \cite{CamachoEtAl2017} evaluates pairs of words from different languages for similarities\footnote{The task was designed to distinguish similarity from relatedness \cite{HillEtAl2015}. An example of the distinction in English-German would be, e.g. \textit{dog-Hund}, which are very similar, whereas \textit{leaf-Baum} would be more dissimilar yet still related.} in their underlying meaning on a scale from 0--4, with a step size of 0.5. 

Given semantic similarity predictions for a list of word pairs constructed per \newcite{CamachoEtAl2015}, performance is measured as the harmonic mean of Pearson and Spearman correlations with human judgments. Datasets were constructed from English, Farsi, German, Italian, and Spanish. As in the UBLI experiments, we train \textit{fastText} embeddings for each language pair using domain-matched (Wiki-Wiki) and domain-mismatched (UN-Wiki) corpora, and compare separate and joint pre-training. Finally, semantic similarity is computed by cosine similarity.

\subsection{Results}

\begin{table}[ht]
\footnotesize
\begin{tabu}{@{}lrrrr@{}}
    \toprule
     & \textbf{En-De} & \textbf{En-Es} & \textbf{En-Fa} & \textbf{En-It} \\
    \midrule 
    Wiki-Wiki & 0.45 & 0.54 & 0.25 & 0.49 \\
    \midrule
    UN-Wiki w/o Joint PT & 0.02 & -0.01 & 0.01 & -0.05 \\
    UN-Wiki w/ Joint PT & 0.43 & 0.47 & 0.23 & 0.45 \\
    $\Delta$ & +0.41 & +0.48 & +0.22 & +0.50 \\
    \bottomrule
\end{tabu}
\caption{Correlation scores for cross-lingual word similarity (SemEval 2017 Task 2, Subtask 2). $\Delta$ refers to the difference between the domain-mismatched scores with and without joint pre-training.}
\label{tab:SevEvalResults}
\end{table}

Results on the cross-lingual semantic word similarity experiments are shown in Table \ref{tab:SevEvalResults}. The Wiki-UN correlation scores are very low without joint pre-training, but the correlation scores improve greatly with joint pre-training and approach scores obtained using matched Wiki-Wiki data, even for a distant language pair like English-Farsi.  


Our experiments on cross-lingual semantic word similarity show that while domain mismatch is a significant obstacle to cross-lingual transfer, joint-training initialization is an effective means of overcoming this issue. Furthermore, unlike the results of the UBLI and UNMT experiments in which joint-training was less effective for the relatively distant language pair of English-Russian in recovering losses compared to matched domain baselines, here joint-training results in virtually identical scores between matched domain and mismatched domain for the distant language pair of English-Farsi.

\section{Conclusion}

Recent publications on UBLI and UNMT have noted that domain mismatch hinders zero-shot cross-lingual transfer \cite{BrauneEtAl2018,tae-unmt-domains,KimEtAl2020}. Our work shows that initialization via joint pre-training can reduce the impact of this mismatch, even when that pre-training doesn't involve post-processing steps such as vocabulary reallocation \cite{WangEtAl2020}. Furthermore, the improvements brought about by this initialization scheme generalize to other cross-lingual tasks such as cross-lingual word similarity. Our results show as much as a 65\% absolute increase in UBLI retrieval accuracy, up to a 21 point gain in UNMT BLEU scores, and as much as a 0.5 improvement on word similarity correlation under domain mismatch. It is well-known that pre-training contextual embeddings on unaligned multilingual corpora induces zero-shot cross-lingual transfer learning (e.g., mBERT and XLM), but our work shows that joint pre-training can also induce simultaneous zero-shot cross-domain, cross-lingual transfer, which we expect will be useful guidance for NLP practitioners.

While the results reported here are encouraging, future work should include experimentation with a wider assortment of language pairs and corpus domains, as well as an investigation of how the distance between languages can affect joint-training's ability to mitigate domain mismatch for different tasks. While in all three tasks it proved very effective for closely related language pairs, the UBLI and UNMT improvements for the more distant language pair of English-Russian were less pronounced. Conversely, on the cross-lingual semantic word similarity experiment, language distance seemed less relevant as joint-training on English-Farsi resulted in scores comparable to the domain matched scenario.

\section{Bibliographical References}\label{reference}

\bibliographystyle{lrec2022-bib}
\bibliography{lrec2022-example}


\end{document}